\pdfoutput=1
\documentclass{opt2023} 

\usepackage[utf8]{inputenc} 
\usepackage[T1]{fontenc}    
\usepackage{hyperref}       
\usepackage{booktabs}       
\usepackage{nicefrac}       
\usepackage{microtype}      
\usepackage{xcolor}         
\usepackage{enumitem}       

\usepackage{csquotes}

\usepackage{varioref}
\usepackage[capitalize]{cleveref}
\usepackage{makecell}

\newcommand{\Mysmall}{\textit{Small}}
\newcommand{\Mybig}{\textit{Medium}}
\newcommand{\Mylarge}{\textit{Large}}

\newcommand{\Myinitial}{\textit{Initial}}
\newcommand{\Mywarm}{\textit{Warm}}
\newcommand{\Mybest}{\textit{Best}}
\newcommand{\Myno}{\textit{No}}

\newcommand{\Myoriginal}{\textit{Original}}
\newcommand{\Myold}{\textit{Old}}
\newcommand{\Myrandom}{\textit{Random}}
\newcommand{\Mytop}{\textit{Top}}

\newcommand{\Myfreezing}{\textit{Freezing}}
\newcommand{\Myidentical}{\textit{Identical}}
\newcommand{\Mydynamic}{\textit{Dynamic}}

\newcounter{defn}[section]
\newenvironment{defn}[1][]{\refstepcounter{defn}\par\medskip
\noindent\textbf{Definition \thedefn: #1} \rmfamily}{\medskip}

\crefname{defn}{\protect{definition}}{\protect{definitions}}
\Crefname{defn}{\protect{Definition}}{\protect{Definitions}}
\creflabelformat{defn}{#2{#1}#3}

\title[Cup Curriculum: Curriculum Learning on Model Capacity]{Cup Curriculum: Curriculum Learning on Model Capacity}

\optauthor{%
\Name{Luca Scharr} \Email{luca.scharr@uni-bonn.de}\\
\Name{Vanessa Toborek} \Email{toborek@cs.uni-bonn.de}\\
\addr University of Bonn, Germany}

\begin{document}

\maketitle

\begin{abstract}
Curriculum learning (CL) aims to increase the performance of a learner on a given task by applying a specialized learning strategy.
This strategy focuses on either the dataset, the task, or the model.
There is little to no work analysing the possibilities to apply CL on the model capacity in natural language processing.
To close this gap, we propose the \emph{cup curriculum}.
In a first phase of training we use a variation of iterative magnitude pruning to reduce model capacity.
These weights are reintroduced in a second phase, resulting in the model capacity to show a cup-shaped curve over the training iterations.
We empirically evaluate different strategies of the cup curriculum and show that it outperforms early stopping reliably while exhibiting a high resilience to overfitting.
\end{abstract}
\begin{keywords}
  Curriculum Learning, Natural Language Processing, Iterative Magnitude Pruning
\end{keywords}
\vspace{5mm}
\begin{figure}[h]
    \centering
    \includegraphics[width=\textwidth]{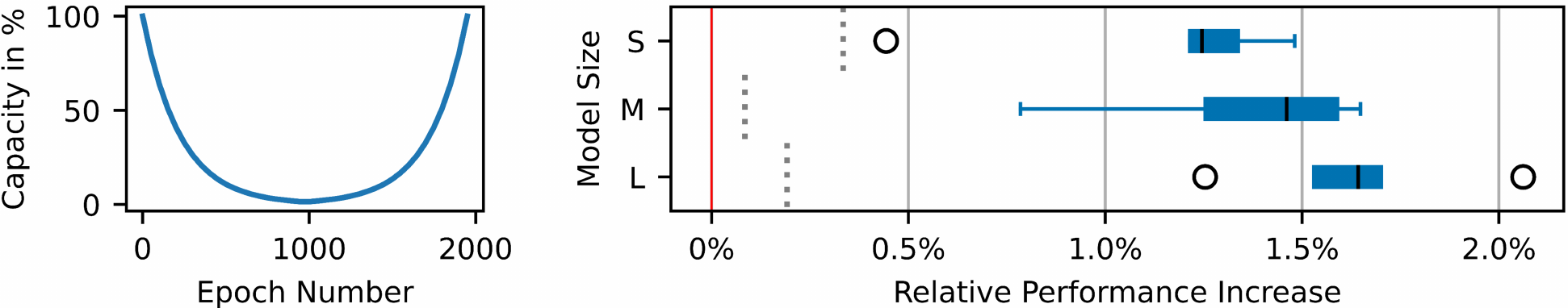}
    \caption{(Left) Model capacity throughout training with the cup curriculum.
    (Right) Relative performance increase of the best performing cup curriculum strategy observed over the best performance of $20$ early stopping runs per model size.
    The red line marks early stopping, the grey line the relative performance increase of the best model found by IMP.}
    \label{fig:FirstPagePlot}
\vspace{-10mm}
\end{figure}
\section{Introduction}
Curriculum Learning (CL) is a well established learning strategy for state of the art approaches in multiple areas of supervised learning including image recognition and natural language processing (NLP) \cite{OHEM,importanceSampling,CLsurvey}.
CL approaches can specify training strategies either for the training data, the task, or the model capacity, where most popular approaches focus on the training data \citep{CLsurvey}.
Although the seminal work of \citet{Elman93} stresses the importance of CL for model capacity, this seems not have been explicitly addressed in literature for NLP so far, especially the reintroduction of capacity.
We therefore in an NLP scenario investigate iterative pruning strategies reducing model capacity followed by iterative reintroduction of model capacity (see \cref{fig:FirstPagePlot} left). 
We formally formulate this as \emph{cup curriculum}. 
In experiments we see that this approach can significantly and reliably improve upon early stopping as well as existing iterative magnitude pruning (IMP) strategies (see \cref{fig:FirstPagePlot} right).

\paragraph{Related Work}\label{Sec:RelatedWork}
It is common knowledge, that children learn exceptionally well.
This is, at least partly, attributed to the large changes in synaptic density in the human brain observed throughout its development in infancy and adolescence \cite{Peter79}.
These changes in synaptic density precisely constitute CL for model capacity.
Therefore, \citet{Elman93} points out the need to apply this variation of CL in the field of NLP.
\citet{Bengio09} introduced CL to computer science but focus on the training data.
Their pioneering work shaped the field of CL in computer science, which largely focuses on the training data \cite{CLsurvey}.
Few work utilize CL for model capacity such as \citet{Morerio17} and \citet{Sinha20}.
The former achieves it by utilising dropout and progressing the dropout probability of neurons.
The latter gradually reduces noise to impair the model capacity during the training. 
However, none of these works are in the field of NLP. 
The idea of pruning parts of a neural network and introducing equivalent parts in terms of expressive capacity later in the training procedure was presented by \citet{RePr}.
However, they studied this idea for Convolutional Neural Networks (CNN) and only for a few filters at a time.

Our training strategy requires the reduction of model capacity, which makes a suitable pruning algorithm paramount.
We decided on a variation of layer wise IMP, as proposed by \citet{DeconstructingLotteryTickets}.
This pruning algorithm is very similar to those used in \citet{LotteryTicket} and \citet{Lin20}.
We decided to stay closer to the former (see \cref{Alg:PruningProcedure}), as it was applied successfully to the transformer architecture before \cite{LotteryTicketsFacebook,LotteryTicketAachen}.
During the second part of our training strategy, the model capacity is increased.
To do so we select the introduction scheme promoted by \citet{RePr} and tailor it for our problem setting (see \cref{Alg:IntroductionProcedure}).

\paragraph{Contributions}
To the best of our knowledge this is the first paper providing an analysis of CL for model capacity in NLP.
We formulate a new CL approach for the model capacity, the cup curriculum, and apply it to the transformer architecture.
The cup curriculum is characterized by two sequential training phases: first, we create a curriculum by reducing the expressiveness of a given model, then we increase the capacity by reintroducing the weights during training.
The transformer architecture is selected for our analysis due to its importance for the field of NLP \cite{Transformer,BERT,GPT3,PaLM,PaLM2,Llama2}.
In our hyperparameter search we were able to find promising cup curricula, which are able to reliably (confidence of $99\%$) improve over the performance reached by early stopping.
Summarizing, we (i) introduce a model curriculum {(\Cref{Sec:CupCurriculum})}, (ii) diligently test different strategies {(\Cref{Sec:Experiments})}, and (iii) provide easy to use code  {(\href{https://github.com/luca-scharr/CupCurriculum}{GitHub})}.

\section{Curriculum Learning for Model Capacity}
We say a model learns from data with respect to some task if its performance as measured by some performance measure improves over time \citep{Mitchell97}. 
CL for model capacity can be characterized by a manipulation of model capacity (measured by some capacity measure) during training with the aim of improving the model's performance \cite{CLsurvey}.

\paragraph{Problem Formulation}
Similar to multiple other machine learning approaches, the goal of our training strategy is the reduction of loss.
This naturally coincides with an improvement in the performance measure used for training, in our case the model's perplexity per word.
Given a task and dataset, our goal is to find a parametrization $\Theta_{\text{CL}}$ of a given model architecture, such that it performs better than the parametrization $\Theta_{\text{No CL}}$ retrieved without CL.
In contrast to $\Theta_{\text{No CL}}$, $\Theta_{\text{CL}}$ may have more, or less, parameters than the initial state $\Theta_0$.
We use IMP to specify the number of model parameters available throughout training. 

\paragraph{Curriculum Generation}
We utilize the iterative nature of IMP to generate a curriculum for the model capacity.
IMP is based on pruning cycles which span a set number of epochs.
After the training in each cycle a pruning criterion is used to prune a percentage of weights individually.
By specifying the number of pruning cycles, training epochs per pruning cycle, and the pruning criterion we receive a curriculum.
Based on the analysis by \citet{DeconstructingLotteryTickets}, we decided to use the magnitude change $c$ as the pruning criterion ($c =  \|w_c\|-\|w_i\|$, with current weight $w_c$ and initial weight $w_i$).
We utilize our cup curriculum framework to refine the received curriculum.

\section{Cup Curriculum} \label{Sec:CupCurriculum}
Utilizing IMP and the insights of \citet{RePr} about the reduction and reintroduction of model capacity, we are able to formulate the \emph{cup curriculum}: 
\begin{defn}[Cup Curriculum] \label{Def:CupCurriculum}
    Let $M$ be a learner, $D$ a dataset, $P$ a performance measure, $C$ a capacity measure, and let $C_M$ denote the capacity of $M$ according to $C$.
    Specify a pruning algorithm, a number of pruning cycles $n$, and a number of growth cycles $m$.
    For each of those cycles specify the number of training epochs $e_i$ and the resulting model capacity $c_i$, where $i \in \{1,\dots,n+m\}$. We also require \mbox{$c_1 > c_2 > \dots > c_n$} and $c_{n+1} < c_{n+2} < \dots < c_{n+m}$.
    
    Together, the above define a \emph{cup curriculum}.
    Training $M$ with it works as follows:
\begin{description}[noitemsep,topsep=5pt]
        \item[Pruning Phase] For $i \in \{1,\dots,n\}$ do:
        \begin{enumerate}[noitemsep,topsep=0pt]
            \item Train $M$ for $e_i$ epochs.
            \item Prune $M$, such that $C_M = c_i$ after completion of the pruning. 
            \item Optionally, rewind the unpruned parameters of $M$ to an earlier state.
        \end{enumerate}
        \item[Growth Phase] For $i \in \{n+1,\dots,n+m\}$ do:
        \begin{enumerate}[noitemsep,topsep=0pt]
            \item Introduce capacity to $M$, such that $C_M = c_i$ after completion of the introduction. 
            \item Train $M$ for $e_i$ epochs.
        \end{enumerate}
        \item[Report] Version of $M$ performing best according to $P$.
    \end{description}
\end{defn}
\Cref{Def:CupCurriculum} specifies a family of curricula for the model capacity.
In our experiments we fix $D$ to WikiText2 \cite{WikiText2}, $P$ to cross-entropy loss, $C$ to percentage of trainable weights, $n$ and $m$ to $20$, $e_i$ to $50$ for $i \in \{1,\dots,n+m\}$, and the pruning algorithm to IMP.
Additionally, we decay $c_i$ by 20\% for $i \in \{1,\dots,n\}$ and set $c_i = c_{n+m-i+1}$ for $i \in \{n+1,\dots,n+m\}$.
We evaluate different rewinding strategies for the pruning phase.
The features extracted during this phase are considered to be essential for the model, as IMP aims to preserve the subspaces of the function space which are most important in the neural network \cite{LotteryTicket}.
Thus, we reintroduce the pruned capacity to the network based on their pruning order (last in first out) aiming to preserve these very features.
We evaluate different initialization schemes which specify the value of reintroduced weights.
Further, we analyze update schemes for the weight update depending on the introduction time of the weight, as further measures may be required to preserve important weights.

Overall, we evaluate the effects of the selected rewinding, initialization, and update scheme on the model performance and generalisation for three different model sizes.
All results are compared to early stopping and among each other.

\paragraph{Model Sizes}
We consider three different model variations of the transformer differing in total parameter count.
They are referred to as the \Mysmall, \Mybig \/, and \Mylarge \/ model (see \cref{Tab:ModelSizes} for the exact number of attention heads and encoder layers).

\begin{table}
\vspace{-.6mm}
\parbox[t][][t]{.4\textwidth}{
\centering
\parbox[t][14mm][t]{0.4\textwidth}{\caption{All rewinding schemes applied during the pruning phase.}}
\label{Tab:RewindingSchemes}
\centering
\begin{tabular}{ll}
    \toprule
    \makecell{Rewinding Scheme} & \makecell{Rewinding State}\\
    \midrule
    \Myinitial & Initial State $\Theta_0$ \\
    \rule{0pt}{10.3pt}\Mywarm & Warm-up State $\Theta_{\text{warm}}$ \\
    \Mybest & Best State $\Theta_{\text{best, cycle}}$ \\
    \Myno & No Rewinding done\\
    \bottomrule
\end{tabular}
}
\hfill
\parbox[t][][t]{.5\textwidth}{
\centering
\parbox[t][14mm][t]{0.5\textwidth}{\caption{All initialization schemes for\\reintroduced weights.}}
\label{Tab:InitializationSchemes}
\begin{tabular}{ll}
    \toprule
    Initialization Scheme & Initialization State\\
    \midrule
    \Myoriginal & Initial State $\Theta_0$ \\
    \Myrandom & Randomised State $\hat{\Theta}$ \\
    \Myold & Pruning State $\Theta_{\text{last, cycle}}$ \\
    \Mytop & Best State $\Theta_{\text{best, cycle}}$ \\
    \bottomrule
    \\
\end{tabular}
}
\vspace{-6.5mm}
\end{table}
\paragraph{Rewinding Schemes}
We test four different rewinding schemes (see \cref{Tab:RewindingSchemes}).
The naive \Myinitial\/ scheme implements the original IMP \cite{LotteryTicket}, which rewinds weights repeatedly to their initial state $\Theta_0$.
In accordance with other results regarding the training of transformer models \cite{LotteryTicketsFacebook,LotteryTicketAachen}, we test the \Mywarm\/ scheme that rewinds weights to a warm-up state $\Theta_{\text{warm}} = \Theta_3$.
We select this model state according to the relative limits given by \citet{StabelizedLotteryTicket}.
The third scheme \Mybest\/ rewinds to the weights of the best performing model of the current cycle $\Theta_{\text{best, cycle}}$.
Lastly, we compare all schemes to a baseline of no rewinding (\Myno\/) with a fixed number of $50$ iterations. 

\paragraph{Initialization Schemes}
The second training phase of the cup curriculum is based on the work of \citet{RePr}, but their findings are not fully applicable to our settings as we prune single weights instead of entire CNN filters.
We thus explore four different \mbox{initialization schemes (see \Cref{Tab:InitializationSchemes})}.
The \Myrandom\/ scheme rewinds the weights by reevaluating the random distribution \mbox{used to initialize $\Theta_0$}.
We also tested initialization schemes based on the pruning criterion $c$ employed during the first training phase ($c =  \|w_c\|-\|w_i\|$) \cite{DeconstructingLotteryTickets}.
The \Myold\/ scheme reintroduces the current weight $w_c$ and the \Myoriginal\/ scheme the initial weight $w_i$.
Both are motivated through their usage in $c$.
Lastly, the \Mytop\/ scheme initializes each reintroduced weight with its value \mbox{in $\Theta_{\text{best, cycle}}$}, i.e. the model which performed best during the pruning cycle which pruned the respective weight.

\paragraph{Update Schemes}
We analyze the effect of three update schemes in the experiments, being the \Myfreezing\/,  \Myidentical\/, and \Mydynamic\/ update scheme.
Here the \Myidentical\/ update scheme denotes the standard backpropagation update.
We focus our analysis on it, as it produces the best results.
For an in depth overview of the update schemes analyzed in the experiments see \cref{App:UpdateSchemes} (or \cref{Tab:UpdateSchemes}).

\section{Experiments} \label{Sec:Experiments}
We trained a transformer architecture \cite{Transformer} on the WikiText2 dataset \cite{WikiText2}.
We ran each experiment with $5$ unique seeds, a fixed learning rate scheduler, and started the learning rate at $5.0$.
The reported results required $1\,080$ wall clock GPU (Nvidia Geforce GTX 1080 Ti $11\,$GB) hours.
\begin{figure}
    \centering
    \includegraphics[width = \textwidth]{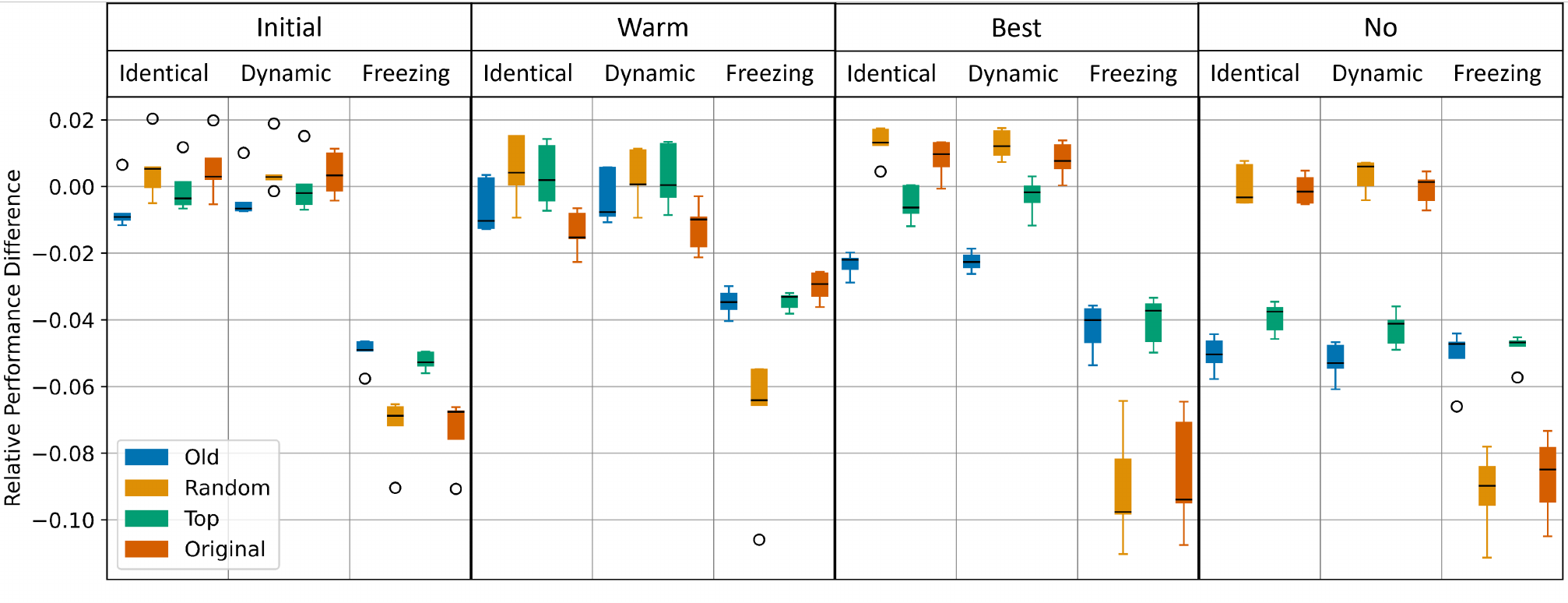}
    \caption{Relative performance difference to the best performing model found by early stopping in 20 runs averaged over all seeds.}
    \vspace{-8mm}
    \label{fig:BigPlotMain}
\end{figure}
\paragraph{Performance}
\Cref{fig:BigPlotMain} shows the relative performance difference of all tested curricula to the best performance of $20$ early stopping runs for the \Mysmall\/ model size.
The best performing strategy is \Mybest\/ rewinding with \Myrandom\/ initialization and \Myidentical\/ updating.
It shows a relative performance increase over early stopping of at least $0.5\%$.
However, we also observe improvements of over $1.5\%$.
\Cref{fig:FirstPagePlot} (right) shows that these improvements increase the larger the model size becomes, even showing an improvement of over $2\%$.
Additionally a median relative \mbox{performance increase of $1\%$ to $1.5\%$} over the best performance of $5$ IMP runs is shown.
All of these improvements are obtained reliably with a confidence of $99\%$ ($\alpha = 0.01$), as measured by the Wilcoxon-Mann-Whitney test \cite{WilcoxonMannWhitney}.
It is surprising, but convenient, that the simplest considered version of the cup curriculum outperforms more complex ones.

Of the results seen in \cref{fig:BigPlotMain}, it is particularly noticeable that strategies which use the \Myidentical\/ and the \Mydynamic\/ update scheme significantly outperform those using the \Myfreezing\/ update.
Furthermore, the \Myrandom\/ and \Myoriginal\/ initialization work best for all rewinding schemes but \Mywarm\/ rewinding, where the \Myrandom\/ and \Mytop\/ initialization work best.
Additionally, for \Mybest\/ and \Myno\/ rewinding the \Myold\/ and \Mytop\/ initialization perform much worse than the other initialization schemes (not counting the \Myfreezing\/ update).
Despite this interaction, no significant difference between the \Myidentical\/ and the \Mydynamic\/ update scheme can be observed for any of the experiments.

As an example of a training curve, \cref{fig:SBestRngIde} shows the training and validation loss of the best performing cup curriculum strategy found in the experiments.
It exhibits resilience to overfitting in addition to the improvements over early stopping and IMP.
While other strategies achieve different performance, all show similar resilience to overfitting (see \cref{fig:OverviewTrainingSmallModel}).
In contrast to early stopping, which prevents overfitting, the cup curriculum could further improve achieved performance with additional training. 
\vspace{-4mm}
\section{Conclusion}
\vspace{-1mm}
In this work we highlight the existence of CL strategies for model capacity in the field of NLP and derive a framework for it: the cup curriculum.
In addition to this proof of concept we analyze multiple strategies of the cup curriculum.
This includes a strategy which reliably (confidence $99\%$) outperforms the best performance found by early stopping over multiple runs by an observed magnitude of $0.5\%$ to $2\%$ while staying resilient to overfitting with the prospect of increasing performance with additional training.
Future work should analyze the impact of different learning rate schedulers on the cup curriculum given the insights of \citet{Llama2} and test our curriculum on state-of-the-art LLMs with the prospect of circumventing the high number of checkpoints and their connected costs as mentioned by \citet{PaLM}.
\paragraph{Acknowledgements}
VTs research has been funded by the Federal Ministry of Education and Research of Germany and the state of North-Rhine Westphalia as part of the Lamarr-Institute for Machine Learning and Artificial Intelligence Lamarr22B.
\bibliography{ref}

\begin{thebibliography}{24}
\providecommand{\natexlab}[1]{#1}
\providecommand{\url}[1]{\texttt{#1}}
\expandafter\ifx\csname urlstyle\endcsname\relax
  \providecommand{\doi}[1]{doi: #1}\else
  \providecommand{\doi}{doi: \begingroup \urlstyle{rm}\Url}\fi

\bibitem[Anil et~al.(2023)Anil, Dai, Firat, Johnson, Lepikhin, Passos, Shakeri, Taropa, Bailey, Chen, Chu, Clark, Shafey, Huang, Meier-Hellstern, Mishra, Moreira, Omernick, Robinson, Ruder, Tay, Xiao, Xu, Zhang, Abrego, Ahn, Austin, Barham, Botha, Bradbury, Brahma, Brooks, Catasta, Cheng, Cherry, Choquette-Choo, Chowdhery, Crepy, Dave, Dehghani, Dev, Devlin, Díaz, Du, Dyer, Feinberg, Feng, Fienber, Freitag, Garcia, Gehrmann, Gonzalez, Gur-Ari, Hand, Hashemi, Hou, Howland, Hu, Hui, Hurwitz, Isard, Ittycheriah, Jagielski, Jia, Kenealy, Krikun, Kudugunta, Lan, Lee, Lee, Li, Li, Li, Li, Li, Lim, Lin, Liu, Liu, Maggioni, Mahendru, Maynez, Misra, Moussalem, Nado, Nham, Ni, Nystrom, Parrish, Pellat, Polacek, Polozov, Pope, Qiao, Reif, Richter, Riley, Ros, Roy, Saeta, Samuel, Shelby, Slone, Smilkov, So, Sohn, Tokumine, Valter, Vasudevan, Vodrahalli, Wang, Wang, Wang, Wang, Wieting, Wu, Xu, Xu, Xue, Yin, Yu, Zhang, Zheng, Zheng, Zhou, Zhou, Petrov, and Wu]{PaLM2}
Rohan Anil, Andrew~M. Dai, Orhan Firat, Melvin Johnson, Dmitry Lepikhin, Alexandre Passos, Siamak Shakeri, Emanuel Taropa, Paige Bailey, Zhifeng Chen, Eric Chu, Jonathan~H. Clark, Laurent~El Shafey, Yanping Huang, Kathy Meier-Hellstern, Gaurav Mishra, Erica Moreira, Mark Omernick, Kevin Robinson, Sebastian Ruder, Yi~Tay, Kefan Xiao, Yuanzhong Xu, Yujing Zhang, Gustavo~Hernandez Abrego, Junwhan Ahn, Jacob Austin, Paul Barham, Jan Botha, James Bradbury, Siddhartha Brahma, Kevin Brooks, Michele Catasta, Yong Cheng, Colin Cherry, Christopher~A. Choquette-Choo, Aakanksha Chowdhery, Clément Crepy, Shachi Dave, Mostafa Dehghani, Sunipa Dev, Jacob Devlin, Mark Díaz, Nan Du, Ethan Dyer, Vlad Feinberg, Fangxiaoyu Feng, Vlad Fienber, Markus Freitag, Xavier Garcia, Sebastian Gehrmann, Lucas Gonzalez, Guy Gur-Ari, Steven Hand, Hadi Hashemi, Le~Hou, Joshua Howland, Andrea Hu, Jeffrey Hui, Jeremy Hurwitz, Michael Isard, Abe Ittycheriah, Matthew Jagielski, Wenhao Jia, Kathleen Kenealy, Maxim Krikun, Sneha Kudugunta, Chang
  Lan, Katherine Lee, Benjamin Lee, Eric Li, Music Li, Wei Li, YaGuang Li, Jian Li, Hyeontaek Lim, Hanzhao Lin, Zhongtao Liu, Frederick Liu, Marcello Maggioni, Aroma Mahendru, Joshua Maynez, Vedant Misra, Maysam Moussalem, Zachary Nado, John Nham, Eric Ni, Andrew Nystrom, Alicia Parrish, Marie Pellat, Martin Polacek, Alex Polozov, Reiner Pope, Siyuan Qiao, Emily Reif, Bryan Richter, Parker Riley, Alex~Castro Ros, Aurko Roy, Brennan Saeta, Rajkumar Samuel, Renee Shelby, Ambrose Slone, Daniel Smilkov, David~R. So, Daniel Sohn, Simon Tokumine, Dasha Valter, Vijay Vasudevan, Kiran Vodrahalli, Xuezhi Wang, Pidong Wang, Zirui Wang, Tao Wang, John Wieting, Yuhuai Wu, Kelvin Xu, Yunhan Xu, Linting Xue, Pengcheng Yin, Jiahui Yu, Qiao Zhang, Steven Zheng, Ce~Zheng, Weikang Zhou, Denny Zhou, Slav Petrov, and Yonghui Wu.
\newblock {PaLM 2 Technical Report}.
\newblock \emph{arXiv: 2305.10403}, 2023.
\newblock URL \url{https://arxiv.org/abs/2305.10403}.

\bibitem[Bengio et~al.(2009)Bengio, Louradour, Collobert, and Weston]{Bengio09}
Yoshua Bengio, J\'{e}r\^{o}me Louradour, Ronan Collobert, and Jason Weston.
\newblock {Curriculum Learning}.
\newblock In \emph{ICML}, page 41–48. Association for Computing Machinery, 2009.
\newblock URL \url{https://doi.org/10.1145/1553374.1553380}.

\bibitem[Brix et~al.(2020)Brix, Bahar, and Ney]{LotteryTicketAachen}
Christopher Brix, Parnia Bahar, and Hermann Ney.
\newblock {Successfully Applying the Stabilized Lottery Ticket Hypothesis to the Transformer Architecture}.
\newblock In \emph{ACL}, pages 3909--3915. Association for Computational Linguistics, 2020.
\newblock URL \url{https://aclanthology.org/2020.acl-main.360}.

\bibitem[Brown et~al.(2020)Brown, Mann, Ryder, Subbiah, Kaplan, Dhariwal, Neelakantan, Shyam, Sastry, Askell, Agarwal, Herbert-Voss, Krueger, Henighan, Child, Ramesh, Ziegler, Wu, Winter, Hesse, Chen, Sigler, Litwin, Gray, Chess, Clark, Berner, McCandlish, Radford, Sutskever, and Amodei]{GPT3}
Tom Brown, Benjamin Mann, Nick Ryder, Melanie Subbiah, Jared~D Kaplan, Prafulla Dhariwal, Arvind Neelakantan, Pranav Shyam, Girish Sastry, Amanda Askell, Sandhini Agarwal, Ariel Herbert-Voss, Gretchen Krueger, Tom Henighan, Rewon Child, Aditya Ramesh, Daniel Ziegler, Jeffrey Wu, Clemens Winter, Chris Hesse, Mark Chen, Eric Sigler, Mateusz Litwin, Scott Gray, Benjamin Chess, Jack Clark, Christopher Berner, Sam McCandlish, Alec Radford, Ilya Sutskever, and Dario Amodei.
\newblock {Language Models are Few-Shot Learners}.
\newblock In \emph{NeurIPS}, pages 1877--1901. Curran Associates, Inc., 2020.
\newblock URL \url{https://proceedings.neurips.cc/paper_files/paper/2020/file/1457c0d6bfcb4967418bfb8ac142f64a-Paper.pdf}.

\bibitem[Chowdhery et~al.(2022)Chowdhery, Narang, Devlin, Bosma, Mishra, Roberts, Barham, Chung, Sutton, Gehrmann, Schuh, Shi, Tsvyashchenko, Maynez, Rao, Barnes, Tay, Shazeer, Prabhakaran, Reif, Du, Hutchinson, Pope, Bradbury, Austin, Isard, Gur-Ari, Yin, Duke, Levskaya, Ghemawat, Dev, Michalewski, Garcia, Misra, Robinson, Fedus, Zhou, Ippolito, Luan, Lim, Zoph, Spiridonov, Sepassi, Dohan, Agrawal, Omernick, Dai, Pillai, Pellat, Lewkowycz, Moreira, Child, Polozov, Lee, Zhou, Wang, Saeta, Diaz, Firat, Catasta, Wei, Meier-Hellstern, Eck, Dean, Petrov, and Fiedel]{PaLM}
Aakanksha Chowdhery, Sharan Narang, Jacob Devlin, Maarten Bosma, Gaurav Mishra, Adam Roberts, Paul Barham, Hyung~Won Chung, Charles Sutton, Sebastian Gehrmann, Parker Schuh, Kensen Shi, Sasha Tsvyashchenko, Joshua Maynez, Abhishek Rao, Parker Barnes, Yi~Tay, Noam Shazeer, Vinodkumar Prabhakaran, Emily Reif, Nan Du, Ben Hutchinson, Reiner Pope, James Bradbury, Jacob Austin, Michael Isard, Guy Gur-Ari, Pengcheng Yin, Toju Duke, Anselm Levskaya, Sanjay Ghemawat, Sunipa Dev, Henryk Michalewski, Xavier Garcia, Vedant Misra, Kevin Robinson, Liam Fedus, Denny Zhou, Daphne Ippolito, David Luan, Hyeontaek Lim, Barret Zoph, Alexander Spiridonov, Ryan Sepassi, David Dohan, Shivani Agrawal, Mark Omernick, Andrew~M. Dai, Thanumalayan~Sankaranarayana Pillai, Marie Pellat, Aitor Lewkowycz, Erica Moreira, Rewon Child, Oleksandr Polozov, Katherine Lee, Zongwei Zhou, Xuezhi Wang, Brennan Saeta, Mark Diaz, Orhan Firat, Michele Catasta, Jason Wei, Kathy Meier-Hellstern, Douglas Eck, Jeff Dean, Slav Petrov, and Noah Fiedel.
\newblock {PaLM: Scaling Language Modeling with Pathways}.
\newblock \emph{CoRR}, 2022.
\newblock URL \url{https://arxiv.org/abs/2204.02311}.

\bibitem[Devlin et~al.(2019)Devlin, Chang, Lee, and Toutanova]{BERT}
Jacob Devlin, Ming{-}Wei Chang, Kenton Lee, and Kristina Toutanova.
\newblock {BERT: Pre-training of Deep Bidirectional Transformers for Language Understanding}.
\newblock In \emph{{NAACL-HLT} {(1)}}, pages 4171--4186. Association for Computational Linguistics, 2019.
\newblock URL \url{https://doi.org/10.18653/v1/n19-1423}.

\bibitem[Elman(1993)]{Elman93}
Jeffrey~L. Elman.
\newblock {Learning and development in neural networks: the importance of starting small}.
\newblock In \emph{Cognition}, pages 71--99. Elsevier, 1993.
\newblock URL \url{https://www.sciencedirect.com/science/article/pii/0010027793900584}.

\bibitem[Frankle and Carbin(2019)]{LotteryTicket}
Jonathan Frankle and Michael Carbin.
\newblock {The Lottery Ticket Hypothesis: Finding Sparse, Trainable Neural Networks}.
\newblock In \emph{ICLR}. OpenReview.net, 2019.
\newblock URL \url{https://openreview.net/pdf?id=rJl-b3RcF7}.

\bibitem[Frankle et~al.(2019)Frankle, Dziugaite, Roy, and Carbin]{StabelizedLotteryTicket}
Jonathan Frankle, Gintare~Karolina Dziugaite, Daniel~M. Roy, and Michael Carbin.
\newblock {The Lottery Ticket Hypothesis at Scale}.
\newblock \emph{arxiv: 1903.01611}, 2019.
\newblock URL \url{http://arxiv.org/abs/1903.01611}.

\bibitem[Huttenlocher(1979)]{Peter79}
Peter~R. Huttenlocher.
\newblock {Synaptic density in human frontal cortex — Developmental changes and effects of aging}.
\newblock In \emph{Brain Research}, pages 195--205, 1979.
\newblock URL \url{https://www.sciencedirect.com/science/article/pii/0006899379903494}.

\bibitem[Lin et~al.(2020)Lin, Stich, Barba, Dmitriev, and Jaggi]{Lin20}
Tao Lin, Sebastian~U. Stich, Luis Barba, Daniil Dmitriev, and Martin Jaggi.
\newblock {Dynamic Model Pruning with Feedback}.
\newblock \emph{arXiv: 2006.07253}, 2020.
\newblock URL \url{https://arxiv.org/abs/2006.07253}.

\bibitem[Mann and Whitney(1947)]{WilcoxonMannWhitney}
H.~B. Mann and D.~R. Whitney.
\newblock {On a Test of Whether one of Two Random Variables is Stochastically Larger than the Other}.
\newblock In \emph{The Annals of Mathematical Statistics}, pages 50--60. Institute of Mathematical Statistics, 1947.
\newblock URL \url{http://www.jstor.org/stable/2236101}.

\bibitem[Merity et~al.(2017)Merity, Xiong, Bradbury, and Socher]{WikiText2}
Stephen Merity, Caiming Xiong, James Bradbury, and Richard Socher.
\newblock {Pointer Sentinel Mixture Models}.
\newblock In \emph{ICLR}. OpenReview.net, 2017.
\newblock URL \url{https://openreview.net/forum?id=Byj72udxe}.

\bibitem[Mitchell(1997)]{Mitchell97}
Tom~M. Mitchell.
\newblock \emph{{Machine learning, International Edition}}.
\newblock McGraw-Hill Series in Computer Science. McGraw-Hill, 1997.
\newblock URL \url{https://www.worldcat.org/oclc/61321007}.

\bibitem[Morerio et~al.(2017)Morerio, Cavazza, Volpi, Vidal, and Murino]{Morerio17}
Pietro Morerio, Jacopo Cavazza, Riccardo Volpi, Rene Vidal, and Vittorio Murino.
\newblock {Curriculum Dropout}.
\newblock In \emph{ICCV}, pages 3564--3572. Proceedings of the IEEE International Conference on Computer Vision, 2017.
\newblock URL \url{https://ieeexplore.ieee.org/document/8237645}.

\bibitem[Prakash et~al.(2019)Prakash, Storer, Florencio, and Zhang]{RePr}
Aaditya Prakash, James Storer, Dinei Florencio, and Cha Zhang.
\newblock {RePr: Improved Training of Convolutional Filters}.
\newblock In \emph{CVPR}, pages 10658--10667. Proceedings of the IEEE/CVF Conference on Computer Vision and Pattern Recognition, 2019.
\newblock URL \url{https://ieeexplore.ieee.org/document/8953322}.

\bibitem[Shrivastava et~al.(2016)Shrivastava, Gupta, and Girshick]{OHEM}
Abhinav Shrivastava, Abhinav Gupta, and Ross Girshick.
\newblock {Training Region-based Object Detectors with Online Hard Example Mining}.
\newblock \emph{arXiv: 1604.03540}, 2016.
\newblock URL \url{https://arxiv.org/abs/1604.03540}.

\bibitem[Sinha et~al.(2020)Sinha, Garg, and Larochelle]{Sinha20}
Samarth Sinha, Animesh Garg, and Hugo Larochelle.
\newblock {Curriculum By Smoothing}.
\newblock In \emph{NeurIPS}, pages 21653--21664. Curran Associates, Inc., 2020.
\newblock URL \url{https://proceedings.neurips.cc/paper_files/paper/2020/file/f6a673f09493afcd8b129a0bcf1cd5bc-Paper.pdf}.

\bibitem[Soviany et~al.(2021)Soviany, Ionescu, Rota, and Sebe]{CLsurvey}
Petru Soviany, Radu~Tudor Ionescu, Paolo Rota, and Nicu Sebe.
\newblock {Curriculum Learning: A Survey}.
\newblock \emph{arXiv: 2101.10382}, 2021.
\newblock URL \url{https://arxiv.org/abs/2101.10382}.

\bibitem[Touvron et~al.(2023)Touvron, Martin, Stone, Albert, Almahairi, Babaei, Bashlykov, Batra, Bhargava, Bhosale, Bikel, Blecher, Ferrer, Chen, Cucurull, Esiobu, Fernandes, Fu, Fu, Fuller, Gao, Goswami, Goyal, Hartshorn, Hosseini, Hou, Inan, Kardas, Kerkez, Khabsa, Kloumann, Korenev, Koura, Lachaux, Lavril, Lee, Liskovich, Lu, Mao, Martinet, Mihaylov, Mishra, Molybog, Nie, Poulton, Reizenstein, Rungta, Saladi, Schelten, Silva, Smith, Subramanian, Tan, Tang, Taylor, Williams, Kuan, Xu, Yan, Zarov, Zhang, Fan, Kambadur, Narang, Rodriguez, Stojnic, Edunov, and Scialom]{Llama2}
Hugo Touvron, Louis Martin, Kevin Stone, Peter Albert, Amjad Almahairi, Yasmine Babaei, Nikolay Bashlykov, Soumya Batra, Prajjwal Bhargava, Shruti Bhosale, Dan Bikel, Lukas Blecher, Cristian~Canton Ferrer, Moya Chen, Guillem Cucurull, David Esiobu, Jude Fernandes, Jeremy Fu, Wenyin Fu, Brian Fuller, Cynthia Gao, Vedanuj Goswami, Naman Goyal, Anthony Hartshorn, Saghar Hosseini, Rui Hou, Hakan Inan, Marcin Kardas, Viktor Kerkez, Madian Khabsa, Isabel Kloumann, Artem Korenev, Punit~Singh Koura, Marie-Anne Lachaux, Thibaut Lavril, Jenya Lee, Diana Liskovich, Yinghai Lu, Yuning Mao, Xavier Martinet, Todor Mihaylov, Pushkar Mishra, Igor Molybog, Yixin Nie, Andrew Poulton, Jeremy Reizenstein, Rashi Rungta, Kalyan Saladi, Alan Schelten, Ruan Silva, Eric~Michael Smith, Ranjan Subramanian, Xiaoqing~Ellen Tan, Binh Tang, Ross Taylor, Adina Williams, Jian~Xiang Kuan, Puxin Xu, Zheng Yan, Iliyan Zarov, Yuchen Zhang, Angela Fan, Melanie Kambadur, Sharan Narang, Aurelien Rodriguez, Robert Stojnic, Sergey Edunov, and Thomas
  Scialom.
\newblock {Llama 2: Open Foundation and Fine-Tuned Chat Models}.
\newblock \emph{arXiv: 2307.09288}, 2023.
\newblock URL \url{https://arxiv.org/abs/2307.09288}.

\bibitem[Vaswani et~al.(2017)Vaswani, Shazeer, Parmar, Uszkoreit, Jones, Gomez, Kaiser, and Polosukhin]{Transformer}
Ashish Vaswani, Noam Shazeer, Niki Parmar, Jakob Uszkoreit, Llion Jones, Aidan~N Gomez, {\L}ukasz Kaiser, and Illia Polosukhin.
\newblock {Attention is All you Need}.
\newblock In \emph{NeurIPS}, pages 5998--6008. Curran Associates, Inc., 2017.
\newblock URL \url{https://proceedings.neurips.cc/paper_files/paper/2017/file/3f5ee243547dee91fbd053c1c4a845aa-Paper.pdf}.

\bibitem[Yu et~al.(2020)Yu, Edunov, Tian, and Morcos]{LotteryTicketsFacebook}
Haonan Yu, Sergey Edunov, Yuandong Tian, and Ari~S. Morcos.
\newblock {Playing the lottery with rewards and multiple languages: lottery tickets in RL and NLP}.
\newblock \emph{arXiv: 1906.02768}, 2020.
\newblock URL \url{https://arxiv.org/abs/1906.02768}.

\bibitem[Zhao and Zhang(2015)]{importanceSampling}
Peilin Zhao and Tong Zhang.
\newblock {Stochastic Optimization with Importance Sampling}.
\newblock \emph{arXiv: 1401.2753}, 2015.
\newblock URL \url{https://arxiv.org/abs/1401.2753}.

\bibitem[Zhou et~al.(2019)Zhou, Lan, Liu, and Yosinski]{DeconstructingLotteryTickets}
Hattie Zhou, Janice Lan, Rosanne Liu, and Jason Yosinski.
\newblock {Deconstructing Lottery Tickets: Zeros, Signs, and the Supermask}.
\newblock In \emph{NeurIPS}. Curran Associates, Inc., 2019.
\newblock URL \url{https://proceedings.neurips.cc/paper_files/paper/2019/file/1113d7a76ffceca1bb350bfe145467c6-Paper.pdf}.

\end{thebibliography}
\newpage
\clearpage
\appendix

\section{Algorithms}
\RestyleAlgo{ruled}
\SetKwComment{Comment}{}{}
The pruning procedure constitutes the first phase of the cup curriculum.
In our experiments it consists of multiple pruning cycles.
Each cycle starts by training the model for a set number of epochs with the possibility of employing early stopping.
After training, a set number of weights is pruned according to their magnitude which is measured by the selected pruning criterion $c$.
Depending on the selected training strategy, the remaining weights are rewound to an earlier state.

The data saved throughout \cref{Alg:PruningProcedure} is included as it is relevant for \cref{Alg:IntroductionProcedure}.
\begin{algorithm2e}[htbp]
\caption{Pruning Procedure}
\label{Alg:PruningProcedure}
\KwData{neural network; training data; evaluation data; number of pruning cycles; percentage to prune; number of epochs; rewinding scheme; pruning version}
\KwResult{For each pruning cycle: mask, best and final state; After all pruning cycles: model state}
    model $\gets$ neural network\;
    list of best states $\gets$ [] \Comment*[r]{[] denotes an empty list}
    list of final states $\gets$ []\;
    list of masks $\gets$ []\;
    \For{cycle \textbf{in} number of pruning cycles}{
        best performance $\gets$ worst possible performance\;
        \For{epoch \textbf{in} number of epochs}{
            state of model $\gets$ train model(training data)\;
            performance $\gets$ evaluate model(evaluation data) \Comment*[r]{evaluates the model}
            \If{performance better than best performance}{
                best performance $\gets$ performance\;
                best state $\gets$ state of model\;
            }
        }
        append best state to list of best states \Comment*[r]{best performing model is stored}
        append state of model to list of final states \Comment*[r]{last reached model is stored}
        \For{weight \textbf{in} weights of the model}{
            $w_c$ $\gets$ current value of the weight\;
            $w_i$ $\gets$ initial value of the weight\;
            pruning criterion of weight $\gets$ $\|w_c\| - \|w_i\|$\;
        }
        model $\gets$ rewind weights(rewinding scheme) \Comment*[r]{rewinds the weights}
        model $\gets$ prune weights(pruning criterion, percentage to prune, pruning version)\;
        mask $\gets$ positions remaining\;
        append mask to list of masks\;
    }
    \textbf{Return} model, list of best states, list of final states, list of masks\;
\end{algorithm2e}

The growth phase is the second phase of the cup curriculum.
In our experiments it consists of multiple introduction steps.
Each one starts by reintroducing capacity, i.e. a number of weights, to the model.
The weight value of each reintroduced weight depends on the introduction scheme chosen.
After this introduction of capacity, the model is trained for a set number of epochs.
The weight update during this training depends on the update scheme chosen.

Just like in \cref{Alg:PruningProcedure} we include data used by \cref{Alg:IntroductionProcedure} to improve clarity.

\begin{algorithm2e}[htbp]
    \caption{Growth Procedure}
    \label{Alg:IntroductionProcedure}
    \KwData{training and evaluation data; model state, list of masks, list of best states and list of final states returned by pruning procedure (\cref{Alg:PruningProcedure}); number of introduction steps; number of epochs; initialization scheme; update scheme;}
    \KwResult{A list containing the best model per capacity}
        model $\gets$ model state\;
        reverse the lists received\;
        list of best states $\gets$ [] \Comment*[r]{[] denotes an empty list}
        \For{step \textbf{in} number of introduction steps}{
            mask $\gets$ list of masks at position step\;
            weights to introduce $\gets$ mask where the model has no active weight\;
            introduce weights(initialization scheme, weights to introduce)\;
            best performance $\gets$ worst possible performance\;
            \For{epoch \textbf{in} number of epochs}{
                state of model $\gets$ train model(training data, update scheme)\;
                performance $\gets$ evaluate model(evaluation data) \Comment*[r]{evaluates the model}
                \If{performance better than best performance}{
                    best performance $\gets$ performance\;
                    best state $\gets$ state of model\;
                }
            }
            append best state to list of best states\;
        }
        \Return list of best states
\end{algorithm2e}
\newpage
\section{Details on Training Curves}
\begin{figure}[htbp]
    \centering
    \includegraphics[height = .3\textheight]{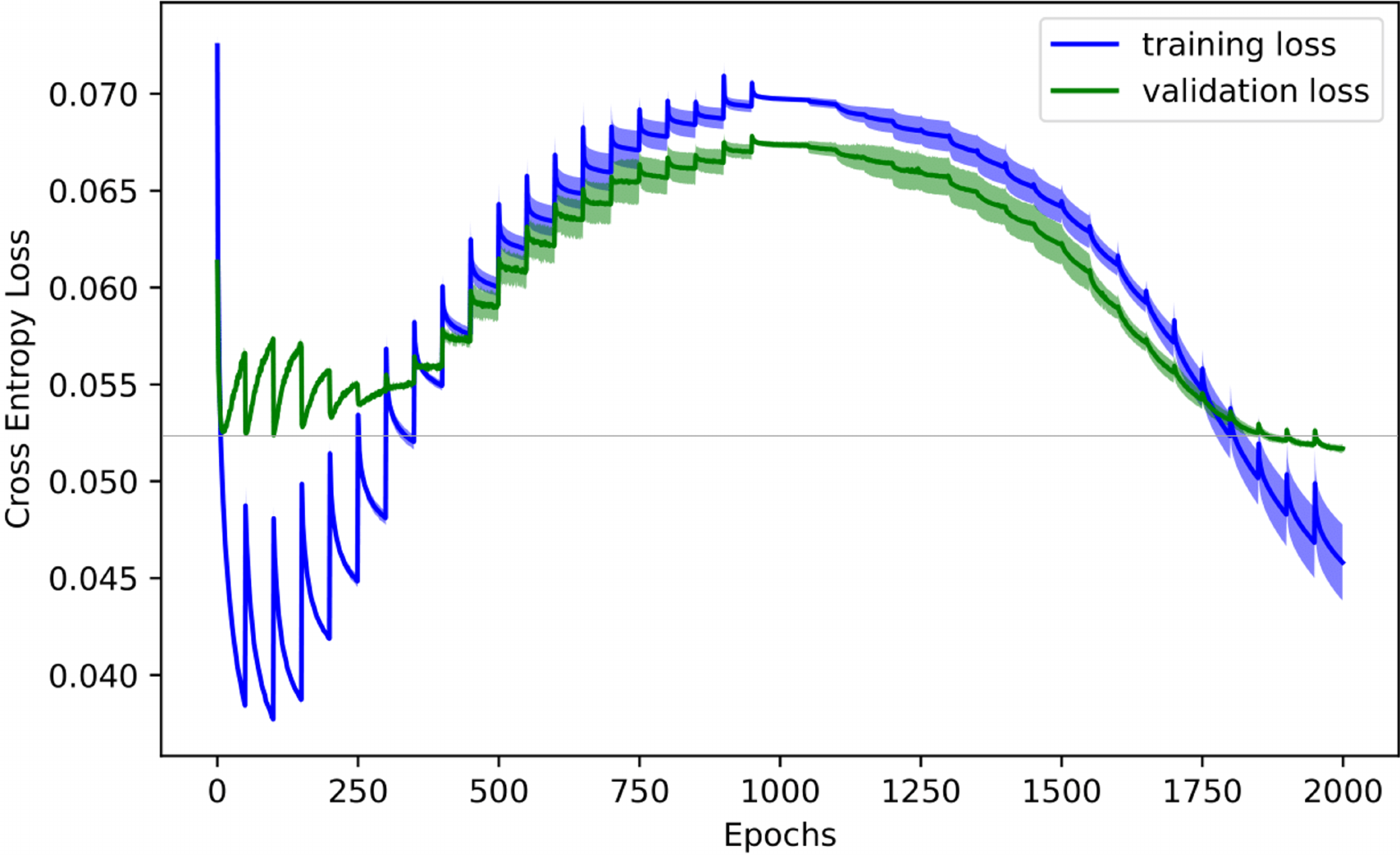}
    \caption{Number of training epochs vs.\ training and validation loss of the cup curriculum strategy using \Mybest\/ rewinding, \Myrandom\/ initialisation, and \Myidentical\/ updating per seed.\\
    The grey line shows the best validation loss of the pruning phase.}
    \label{fig:SBestRngIde}
    \vspace{-5mm}
\end{figure}

\begin{figure}[htbp]
    \centering
    \includegraphics[width = \textwidth]{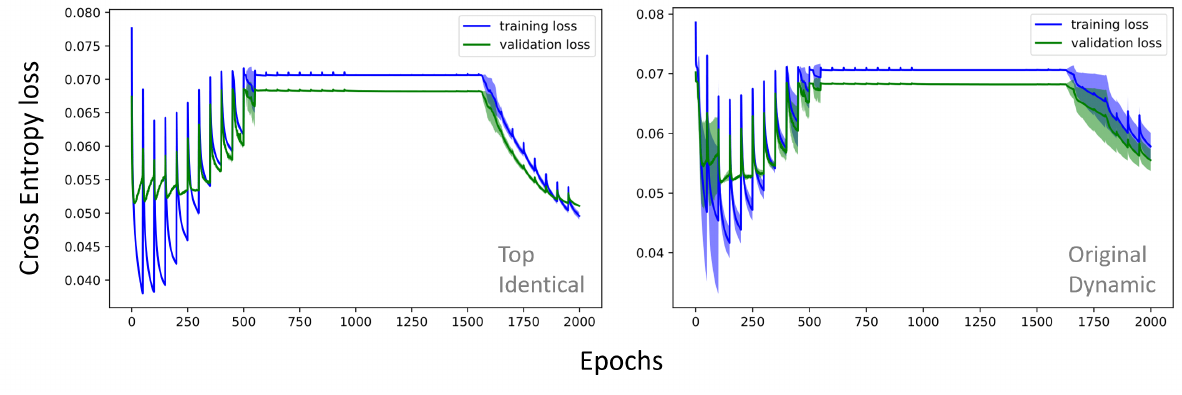}
    \caption{Exemplary training curves for the medium (left) and large (right) model.}
    \label{fig:OverviewTrainingMediumLarge}
\end{figure}

\begin{figure}[htbp]
    \centering
    \includegraphics[width = \textwidth]{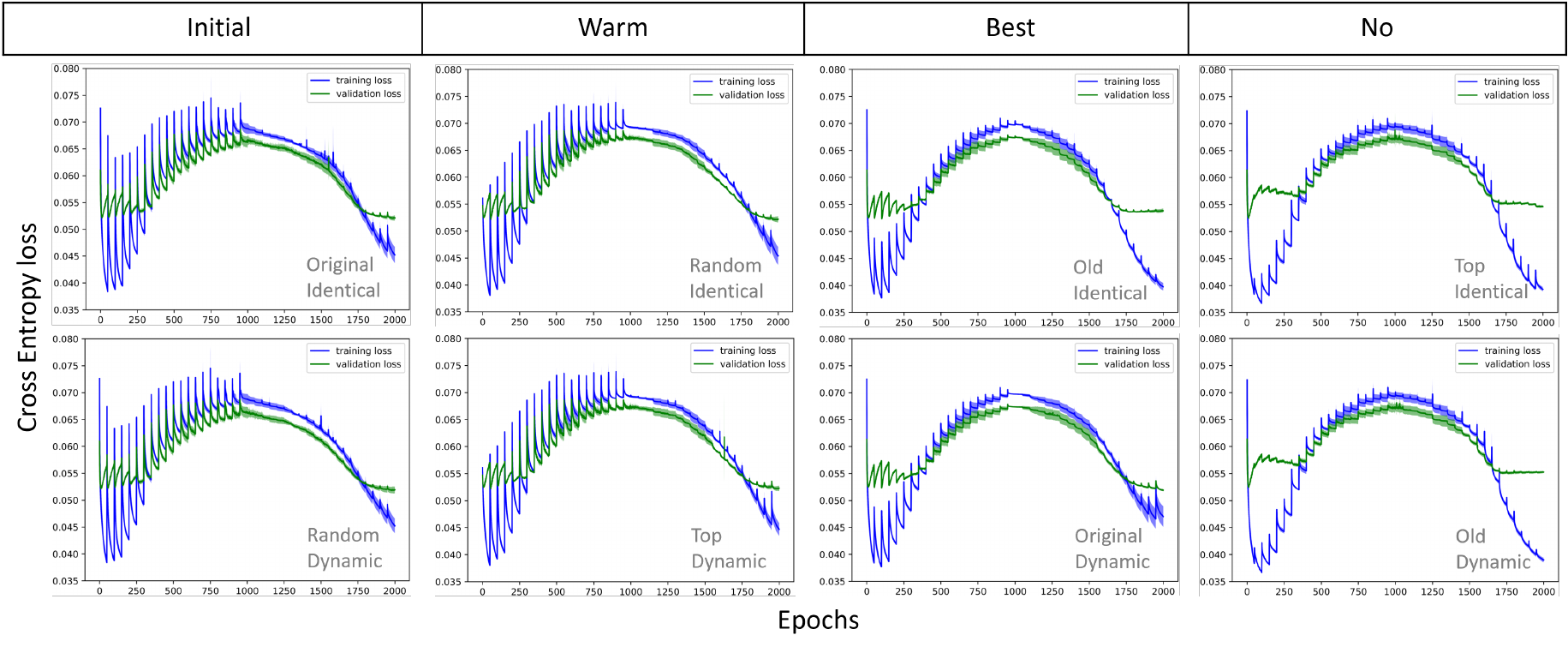}
    \caption{Overview over different training curves of different strategies of the cup curriculum on the small model.}
    \label{fig:OverviewTrainingSmallModel}
\end{figure}

Throughout the pruning phase of each considered training scheme regular spikes in training and validation loss can be observed (see \cref{fig:SBestRngIde,fig:OverviewTrainingMediumLarge,fig:OverviewTrainingSmallModel}).
These spikes occur every $50$ epochs and are due to weight rewinding and subsequent pruning.
Their magnitude depends on the rewinding strategy used.
\Myinitial\/ and \Mywarm\/ rewinding results in larger spikes, as their rewinding state is separated from the current state by more training epochs than the rewinding state of \Mybest\/ and \Myno\/ rewinding.
Similar spikes are present in the growth phase due to the reintroduction of weights.
However, here they are less pronounced.

Due to the usage of dropout throughout the training procedure the model capacity used to measure the training and validation loss differ ($80\%$ compared to $100\%$).
This relative difference in model capacity is our explanation for the validation loss being substantially smaller than the training loss at smaller model capacities (see \cref{fig:SBestRngIde,fig:OverviewTrainingMediumLarge,fig:OverviewTrainingSmallModel}).

For the \Mysmall\/ model size all initialization schemes but the \Myold\/ scheme show resilience to overfitting across multiple initialization and update scheme combinations (see \cref{fig:OverviewTrainingSmallModel}).
Our explanation for this circumstance is the unaddressed overfitting throughout the early pruning cycles of training schemes utilizing the \Myold\/ scheme.
For the \Mysmall\/ model size the training and validation loss of strategies using the other initialization schemes in combination with the \Myidentical\/ or \Mydynamic\/ update scheme are shaped like a bell.
This differs when changing the model size.
For the \Mybig\/ and \Mylarge\/ model sizes the loss plateaus after the $11$-th pruning cycle and only decreases with the beginning of the $11$-th and $12$-th introduction step (see \cref{fig:OverviewTrainingMediumLarge}).
All initialization schemes but the \Myno\/ scheme show resilience to overfitting across multiple rewinding and update scheme combinations.

\newpage

\section{Implementation Details}
Here the remaining parameters mentioned in the paper are explained in more detail.
\begin{table}[h]
\parbox[t][5cm][t]{0.5\textwidth}{
\centering
\parbox[t][14mm][t]{0.5\textwidth}{\caption{Transformer architecture details by model variation.}\label{Tab:ModelSizes}}
\begin{tabular}{lccc}
    \toprule
    Model & \makecell{Attention\\Heads} & \makecell{Encoder\\Layers} & Parameter Count\\
    \midrule
    \Mysmall & 2 & 2 & 13\,829\,280 \\
    \Mybig & 4 & 4 & 14\,313\,280\\
    \Mylarge & 8 & 8 & 15\,281\,280\\
    \bottomrule
    \\
\end{tabular}
}
\hfill
\parbox[t][5cm][t]{0.45\textwidth}{
\centering
\parbox[t][14.4mm][t]{0.4\textwidth}{\caption{All update schemes applied during the growth phase.}\label{Tab:UpdateSchemes}}
\begin{tabular}{lr}
    \toprule
    \makecell{Update\\Scheme} & Weight Update\\
    \midrule
    \makecell[l]{\Myfreezing\\\hfill} & \makecell[r]{only last introduced\\weights are updated}\\
    \Myidentical & same LR for all weights\\
    \makecell[l]{\Mydynamic\\\hfill} & \makecell[r]{LR depends on\\introduction time}\\
    \bottomrule
\end{tabular}
}
\end{table}
\paragraph{Model Sizes}\label{App:ModelSizes}
We consider three model sizes in the experiments to explore the impact of parameter count, number of attention heads, number of encoder layers and dimensionality of attention heads on the training strategies considered.
The \Mysmall\/ model consists of $2$ attention heads and $2$ encoder layers.
The \Mybig\/ model consists of $4$ attention heads and $4$ encoder layers.
Lastly, the \Mylarge\/ model consists of $8$ attention heads and $8$ encoder layers.
The dimension of the projection computed in each attention head scales inversely with the number of attention heads, therefore these model sizes also explore this effect on the training strategies considered.
\Cref{Tab:ModelSizes} lists the considered model sizes.
\paragraph{Update Schemes}\label{App:UpdateSchemes}
In the experiments we analyze the effect of three different update schemes.
First, the \Myfreezing\/ update scheme is considered.
It freezes all but the last introduced weights of the model.
Additionally, we consider the usual weight update (\Myidentical\/) which is plain backpropagation.
Lastly, a weight update based on the individual weights age in the network (\Mydynamic\/) is considered.
The \Mydynamic\/ update scheme multiplies the weight update for all weights introduced in the ${n}$-th introduction step with ${f^{n}}$, where $f \in \mathbb{R}^+_0$ and weights which were not pruned during the pruning phase are considered to be from the $0$-th introduction step.
This extends the idea of freezing the update (by applying update masks with elements in $\{0;1\}$), to update masks with elements in $\mathbb{R}^+_0$.
The update schemes analyzed in the experiments are listed in \cref{Tab:UpdateSchemes}.

\end{document}